# i-GSI: A Fast and Reliable Grasp-type Switching Interface based on Augmented Reality and Eye-tracking*

Chunyuan Shi, Dapeng Yang, *Senior, IEEE*, Siyang Qiu and Jingdong Zhao, *Member, IEEE*

*Abstract*— **The control of multi-fingered dexterous prosthetics hand remains challenging due to the lack of an intuitive and efficient Grasp-type Switching Interface (GSI). We propose a new GSI (i-GSI) t hat integrates the manifold power of eye-tracking and augmentced reality technologies to solve this problem. It runs entirely in a HoloLens2 helmet, where users can glance at icons on the holographic interface to switch between six daily grasp types quickly. Compared to traditional GSIs (FSM-based, PR-based, and mobile APP-based), i-GSI achieved the best results in the experiment with eight healthy subjects, achieving a switching time of 0.84 s, a switching success rate of 99.0 ± 0.6%, and learning efficiency of 93.50%. By verifying on one patient with a congenital upper limb deficiency, i-GSI achieved an equivalent great outcome as on healthy people, with a switching time of 0.78 s and switching success rate of 100%. The new i-GSI, as a standalone module, can be combined with traditional proportional myoelectric control to form a hybrid-controlled prosthetic system that can help patients accomplish dexterous operations in various daily-life activities.**

I. INTRODUCTION

Humans rely heavily on their hands to perform normal daily activities [1]. Patients with hand amputation face many difficulties in life [2]. Wearing a prosthetic hand is the main method to reinstate [3]. Many advanced prosthetics have a good balance of both functionality and appearance. With multiple controllable degrees of freedom (DOFs), these hands [4-9] are generally equipped with independently-moving fingers and can reproduce many daily grasp types [10] and other hand functions. However, the lack of a good human-machine interface (HMI) is the weak point for these prosthetics, congesting the development of many useful functionalities [11].

In real-life applications, it is unrealistic and unnecessary to control every single finger or joint of the hand explicitly. Instead, defining a set of representative grasp types for the specific application that ensures adequate performance will be sufficient [12]. Therefore, how to switch between different grasp types is the key to developing good control. Furthermore, a robust, reliable, and intuitive grasp-type switching interface (GSI) is a fundamental requirement to support dexterous control [11].

For commercial prosthetics, Finite State Machine-based GSI (FSM-based GSI) is commonly used for grasp-type switching [13]. It can switch between three to four grasp types reliably. Still, the user must perform several co-contractions to complete a single task [14], which is counterintuitive and burdensome. The situation gets worse with more grasp types. Bandwidth between humans and prosthetics is insufficient with only two to three electrodes. The same problem exists in the coding-based control method [15].

More electrodes and better algorithms are the most intuitive ways to solve the insufficient bandwidth problem [16]. In research fields, pairing multi-channel (more than six) EMGs with pattern recognition (PR-based GSI) is commonly used to switch between multiple grasp types intuitively and rapidly [17]. However, since these GSIs are sensitive to confounding factors, the outcome in clinical applications is unsatisfactory. Even though more electrodes can improve the robustness, the loss of muscles due to amputation makes EMG signals hardly accessible [14]. Targeted Muscle Reinnervation (TMR) [18] is a solution to muscle loss. Still, it has some risks associated due to the invasive surgery required, such as phantom limb pain, development of painful neuromas [14], etc. On the other hand, multi-DOF simultaneous control methods based on non-negative matrix factorization, deep learning, etc., can achieve synchronous decoding with multiple DOFs [19]–[21]. However, the DOFs are generally limited to two to three, so it is not applicable over multiple grasp types. In addition, whether using the PR-based method or the simultaneous control method, the system's robustness remains a bottleneck, especially when multi-channel EMGs are employed [22].

To overcome the drawbacks of EMG-only control systems and` TMR, hybrid myoelectric control systems using two or more control signals were developed [14]. In these hybrid systems, the proportional myoelectric signal (with two to three channels) [23] is responsible for intuitive control of the opening and closing motion of the prosthesis. Other signals are in charge of the grasp-type switching task to improve the HMI bandwidth. Mobile application (APP), mechatronic switcher, inertial measurement units (IMU), and radio-frequency identification (RFID) tags [24] are existing commercial solutions (e.g., i-limb hand [8]). However, the operation is still complicated and inconvenient. Although the plantar pressure signal [25] and the tongue movement signal [26] improve user experience, the grasp types are insufficient and have poor real-time performance. Using the visual signal is a newly rising method [27]. However, only simple tests have been conducted, and many issues remain to be solved. For one, it is difficult to identify objects and distinguish the object

* This work is partially supported by NSF Grant #52075114 and Postdoctoral Scientific Research Development Fun (LBH-W18058) to D.Yang. Corresponding author: Dapeng Yang (yangdapeng@hit.edu.cn).

The authors are from the State Key Laboratory of Robotics and System, Harbin Institute of Technology (HIT), Harbin 150080, China. D. Yang and J. Zhao are also from the Artificial Intelligence Laboratory (HIT), Harbin 150001, China (e-mail: Chunyuan_Shi, yangdapeng, zhaojingdong@hit.edu.cn, siyangqiu@stu.hit.edu.cn).

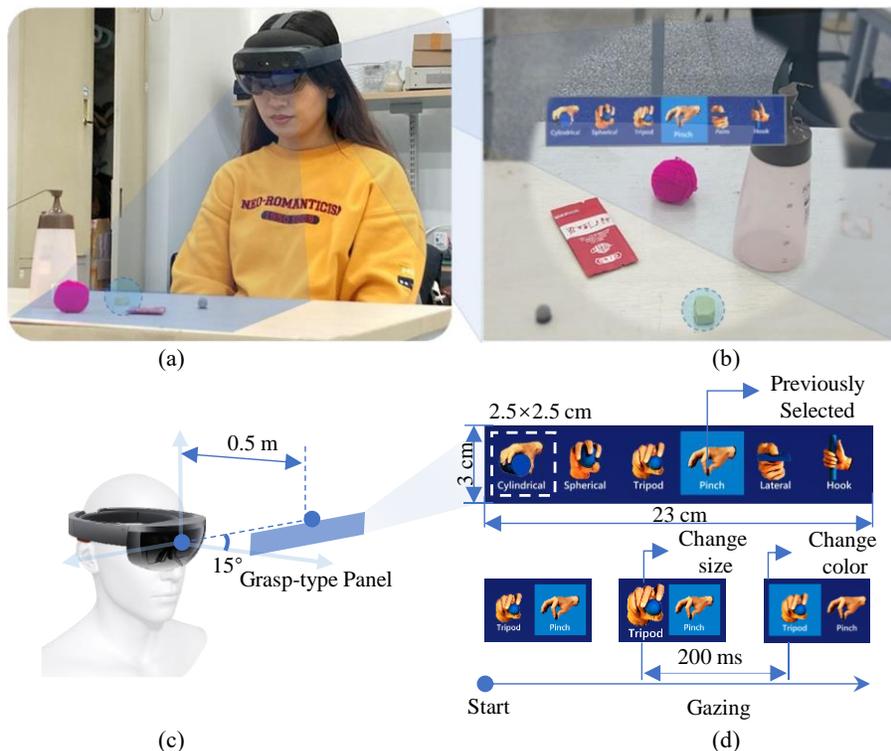

Fig. 1. i-GSI system schematic. (a) Actual scene with the user only wearing HoloLens2. (b) First-person perspective with the pinch grasp type selected to grab the eraser. (c) Display region of panel, within optimal eye movement area and do not block natural sight. (d) Configuration on grasp-type panel and the switching process. A total of six grasp type menus (2.5 cm × 2.5 cm). Select a grasp type by staring at the icon for more than 200 ms.

of interest in cluttered environments [14]. For another, it is very hard for users to intervene in the decision-making process of autonomous systems.

The low performance of GSIs directly limited the effectiveness of hybrid systems. Intuitively, the better the quality of the input signal, the better the performance of the GSIs. Eye movement signal is a stable, easy-to-control, natural, and intuitive signal which contains information regarding the object of interest to the user [28]. The feasibility of using this signal for interactive control has been proven in the research on companion robots [29]. In [29], a paralyzed patient stared at a food image on the monitor to make the robot deliver that corresponding food to his mouth.

Intuitive feedback can greatly improve control precision. Head-mounted augmented reality (AR) devices, such as HoloLens 2 (Microsoft), can naturally superimpose holographic virtual objects in the sight of the user, which can communicate information and action feedback to the user. Driven by the metaverse concept, AR technology is under rapid development. Many hardware portable products (such as HTC Vive Eye Pro，Magic Leap，Google glass, etc.) could potentially form an integrated system with the prosthetics.

In an early study, we explored using a simple platform (desktop eye-tracking device and screen-based HMI) to validate whether AR and eye-tracking technology have the potential to provide new means for prosthetic hand control [30]. This paper proposes a new GSI combining AR and eye-tracking technology, called i-GSI. Running in Hololens2, i-GSI can be easily integrated into existing dexterous prosthetic systems as a standalone module. We verify the superiority of i-GSI by designing and conducting a comparison experiment between i-GSI and three traditional GSIs (PR-based GSI, FSM-based GSI, and APP-based GSI). Also, we confirm the i-GSI suitability for patients (with weak EMG signals).

## II. MATERIALS AND METHOD

Six daily grasp types were selected from [10] as targets of the GSIs, as shown in TABLE I. These grasp types can cope with most situations in daily life. To evaluate the performance of i-GSI, three traditional GSIs are selected for comparison: PR-based, FSM-based, and APP-based. PR-based GSI is the mainstream researched method. FSM-based GSI is the relatively stable control strategy currently dominating the commercial prosthesis. APP-based GSI is the auxiliary solution of the commercial prosthesis, used as an extension of the FSM-based GSI (e.g., I-limb hand [8]).

### A. i-GSI

i-GSI realizes the switching of grasp types by detecting gaze behaviour on the virtual grasp-type icons. To ensure the user can obtain real-time grasp-type feedback, i-GSI generates a grasp-type panel (Figure 1(d)) and renders it superimposed in front of the user (Figure 1(c)). To detect the grasp type of interest, i-GSI obtains the gaze direction to generate a virtual sight. A gaze behaviour is triggered when the virtual sight collides with a grasp-type icon and remains for more than a gaze duration threshold. To avoid being over-sensitive and falsely triggered, the i-GSI gaze duration threshold was set to 200 ms (generally, lasting 120 ms is considered a gaze behaviour). When the user glances at a specific grasp-type icon, i-GSI will select the grasp type and send relevant commands to outer devices through wireless communication. In the paper, it is sent to the PC through WIFI for experimentation. But in real scenarios, it can be transmitted to commercial prosthetics (e.g., i-limb) through WIFI or Bluetooth.

For future convenience of integrating into existing prosthetics, i-GSI is modularly compiled into an independent application implemented using the Unity engine and C# language, which runs entirely on an AR helmet. The helmet, HoloLens 2 from Microsoft (Figure 1(a)), allows users to simultaneously see reality and holographic virtual objects (Figure 1(b)) while detecting eye movement information with the built-in eye tracker.

*B. PR-based GSI*

The commercial MYO armband (eight channels, annular arrangement) [31] was selected as the EMG electrodes. The built-in pattern recognition algorithm can recognize five myoelectric patterns (see TABLE I). To realize the PR-based GSI, the five EMG patterns are one-to-one, corresponding to the first five grasp types (TABLE I) of the grasp-type panel (Figure 1(d)). That is, when the user makes an EMG pattern, the corresponding grasp type will be selected. The algorithm of this GSI is implemented in Python, and the MYO is connected to the PC via Bluetooth. For fairness, the GSI integrated the same feedback as the i-GSI.

TABLE I. EMG PATTERN – GRASP TYPE MATCHUP

| EMG Patterns | | Grasp Type |
|---|---|---|
| Wave In |  | Cylindrical |
| Wave Out |  | Spherical |
| Fist |  | Tripod |
| Fingers Spread |  | Pinch |
| Double Tap |  | Lateral |
| - | - | Hook |

*C. FSM-based GSI*

FSM-based GSI generally sets each grasp type as a different state and allocates a certain EMG pattern as a trigger signal to switch states in sequence. This paper uses the *Fist* EMG pattern described in B as the switching signal (Figure 2). This GSI also integrated AR functionality as an addition.

*D. APP-based GSI*

APP-based GSI requires the patient to use their healthy hand to tap on mobile phone or tablet buttons to switch the grasp type. A mobile phone (Apple iPhone 11, screen size: 6.1") was selected, and a customized program was installed on the device (Figure 3). Since prompts exist on the APP interface, no AR feedback is utilized. In the following experiments, the mobile device was set to always-on mode, so the subjects did not need to wake up the device.

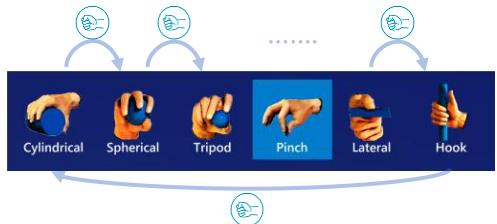

Fig. 2. FSM-based GSI illustration

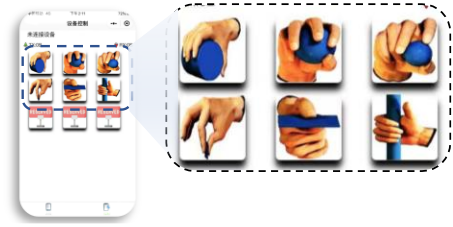

Fig. 3. Mobile APP user interface

### III. EXPERIMENTS

A GSI comparison experiment was designed, and eight healthy subjects were recruited to compare the performance of different GSIs. In addition, one patient was also recruited to verify the patient adaptability of i-GSI.

*A. Experimental Protocols*

The subjects can be in the standby or operation phase in the experimental scene (Figure 4(a)). In the standby phase, the subjects face away from the table. They cannot see the target object on the table and have no clue what grasp type they should use. In the operation phase, they turn to the table. They can see the object and freely switch the grasp type. An IMU detects postures representing each phase in HoloLens2.

*B. Task Flow*

The task-flow diagram is shown in Figure 4(b). The assistant will place the target object in area b and give instructions to the subject in the standby phase. After instructed, the subject will adjust to the operation phase (the timer starts automatically), observe the target object, and use the GSI to switch to the target grasp type. In the end, the subject faces away from the desk and returns to the standby phase (the timer stops automatically).

*C. Metrics*

To evaluate the switching speed and accuracy of the GSIs, two indicators, switching time (ST) and switching success rate (SSR), are defined. ST refers to the time from the subject initially putting their eyes on the object to selecting the correct grasp type at last. SSR refers to the proportion of correct grasp types selected after completing all tests. The Learning Efficiency (LE) indicator is defined to evaluate the mastery efficiency of different GSIs. This indicator is calculated based on the BCG Experience Curve, which the formula is shown below:

$$y = kx^n \tag{1}$$

$$n = {lgb}/{lg2} \tag{2}$$

where $x$ is the test index, $y$ is the time taken on the $x^{th}$ test, and $b$ is learning Efficiency (LE).

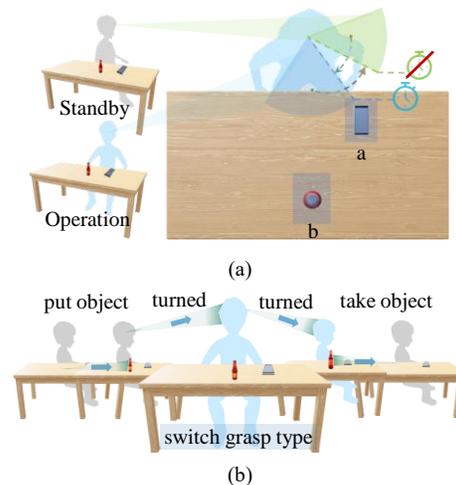

Fig. 4. Experimental scene illustration. (a) Experiment setup. a mobile device in region a (if applicable) and target object in region b. (b)Task flow diagram.

### D. Subjects

Eight healthy subjects (seven males, one female, age 23.5 ± 2.5, height 175 ± 8 cm, weight 68 ± 11 kg) were recruited to participate in the comparative experiment of four GSIs. All subjects had no prior experience with myoelectric control and prosthetics usage. In addition, a patient with congenital upper limb deficiency (male, age 26, height 178 cm, weight 62 kg, previously participated in prosthetic operation experiment) was recruited. This patient only attempted the i-GSI method to explore the suitability of the i-GSI on patients (result in the Discussion section). All experiments were approved by the university's Ethical Committee (NO.HIT-2021009) and conformed to the Declaration of Helsinki.

### E. Experiment Procedures

Subjects need to be fully inexperienced with any GSI to assess the LE of each GSI. Hence, the subjects were equally divided into four pairs, with each pair allocated to a different GSI (first pair: i-GSI, second pair: PR, third pair: FSM, fourth pair: APP). Each pair start by attempting their allocated GSI for LE evaluation. Then, they attempt the other three GSIs in turn to assess ST and SSR (e.g., the first pair will attempt: i-GSI→PR→FSM→APP).

Each GSI requires 30 sets of consecutive experiments (see part F for more detail) without contaminating other GSIs between the sets. The system sensor was calibrated for each participant before conducting the experiments. The participants took five minutes of rest after each set to prevent fatigue. Out of the 30 sets, the first 20 were regarded as the practice stage to ensure the subjects could master the GSI. The final 10 sets were regarded as the tryout stage, where the experimental results were used for performance evaluation.

### F. 21 Objects and 30 Sets of Sequences

Twenty-one everyday objects were selected to evaluate the GSIs as realistically as possible. Each object can be grasped using one of the six grasp types, including one object for *Hook* and four for each other grasp type (see TABLE II for more detail). The objects are indexed, and 30 sets of random placement sequences are pre-generated using a software program to unify placement sequences for all GSIs. Since only one object for *Hook* and four for each other grasp type, the object for *Hook* appears four times. Hence, there are "24 objects" to be grasped in each sequence. Another constraint is that neighbouring objects in each sequence should require different grasp types. Since FSM is a sequential switching method, the ST is directly proportional to the number of co-contractions (NCCs) required, varying between one to five NCCs. Great efforts were made to guarantee that every grasp-type switching situation appears with an equal probability.

TABLE II. 21 TARGET OBJECTS

| Grasps Types | IMG | Objects |
|---|---|---|
| Cylindrical | 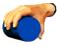 | 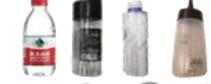 |
| Spherical |  | 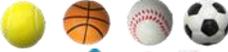 |
| Tripod |  | 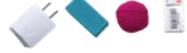 |
| Pinch | 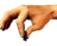 | 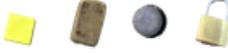 |
| Lateral |  | 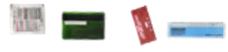 |
| Hook | 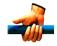 | 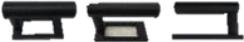 |

### G. Sensor Calibration

We calibrated the sensor for the experimenter before they started the experiment. The angle threshold of the standby phase and operation phase was calibrated. The eye movement calibration application in HoloLens2 was used to calibrate the eye movements of each subject. According to the product specifications, the MYO was worn at the top one-third position on the right forearm (near the elbow) and calibrated for the five EMG patterns with the software application from MYO. All subjects were instructed to put their left-hand flat on the left side of the mobile phone and only operate with the left hand (any finger) during the test.

### H. Data Analysis Methods

LE was obtained by fitting ST data into the BCG Experience Curve (1) and substituting it in (2). The ST of each GSI was tested for normal distribution. Unfortunately, not even one was normally distributed. Therefore, boxplots and medians were used. The nonparametric rank-sum test (Kruskal-Wallis test) and Bonferroni correction were used for significance testing.

## VI. RESULTS

The eight healthy subjects completed 22080 tests: 14720 tests were performed in the practice stage (first 20 sets), and 7360 tests were in the tryout stage (final 10 sets).

### A. Learning Efficiency

As described in Section II-C, LE is analyzed using the 30 sets of experimental data from the first GSI attempted by each subject. The results in Figure 5 show that the new i-GSI has the best LE, reaching 93.50%. PR-based GSI is the lowest, with only 81.50%. The other two lay in the middle, with APP-based GSI at 90.00% and FSM-based GSI at 86.90%. The result above is consistent with the operational complexity of the GSIs:

1) i-GSI: switch by staring at the suitable grasp-type icon (simple).

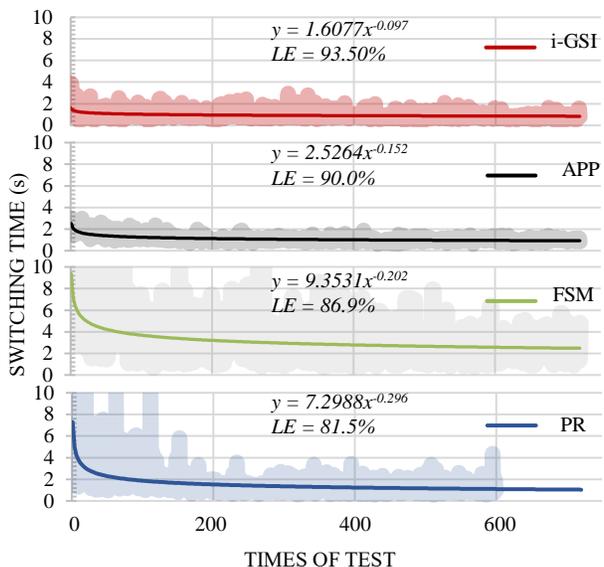

Fig. 5. BCG experience curve in four methods

2) <u>APP-based GSI</u>: look for the most suitable grasp type on the screen, then tap on the icon to select (simple-medium).
3) <u>FSM-based GSI</u>: trigger the same EMG signal at different times to select different grasp types (medium).
4) <u>PR-based GSI</u>: trigger different EMG signals to select different grasp types (complex).

### B. Proficiency

To ensure that the subjects are proficient in each GSI during the tryout stage, variation in ST distribution of the tryout stage was studied for each GSI: i-GSI was quite indifferent ($P_{\text{i-GSI}} = 0.83$); PR-based GSI showed slight variance ($P_{\text{PR}} = 0.03$) between certain pairs of sets ($P_{\text{PR(22 VS 30)}} = 0.01$, $P_{\text{PR(other pairs)}} > 0.20$); the same goes with APP-based GSI ($P_{\text{APP}} = 0.013$, $P_{\text{APP(25 VS 27)}} = 0.02$, $P_{\text{APP(other pairs)}} > 0.10$); FSM-based GSI, on the other hand, demonstrated significant difference ($P_{\text{FSM}} < 0.01$). From our investigation, the difference mainly occurs between pairs of sets that have large contrast in the total NCCs (NCCs: $N_{\text{Set 21}} = 455$, $N_{\text{Set 22}} = 453$, $N_{\text{Set 24}} = 491$, $N_{\text{Set 25}} = 602$, $N_{\text{Set 27}} = 606$, $520 \leq N_{\text{Others}} \leq 578$). But since the total NCCs are determined entirely by the object placement sequence as detailed in Part II-F, the difference observed here is not caused by a lack of proficiency. All in all, the experimental data of all four GSIs are representative because the subjects were proficient.

### C. Performance

An analysis is made at the overall level and individual levels to compare the performance of the four GSIs:

*1) Overall:* Experimental data of all subjects were collectively studied to analyze the distribution of ST among the four GSIs. The results suggested a significant variance between each pair of GSIs ($P < 0.001$). From the ST results in Figure 6(a), i-GSI is the fastest (0.84 s), FSM-based GSI is the slowest (2.53 s), and the other two are slightly more than 1 s. In terms of the SSR results in Figure 6(b), all GSIs achieved over 98.9%, rather indifferent due to timely feedback. There are only small probability error events dominantly concentrated at the end of the tests.

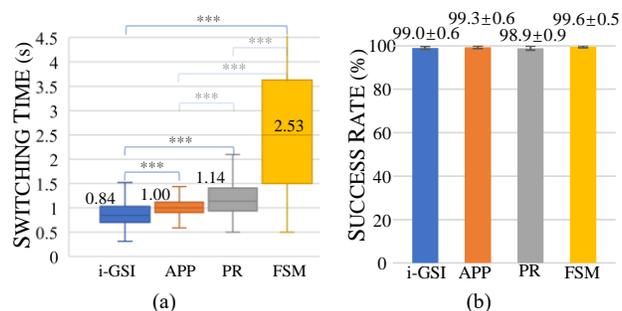

Fig. 6. Result in diff. GSIs ($P_{***} < 0.001$)

*2) Individual:* An additional study on ST distribution over the four GSIs was performed to verify that the overall results represent each subject. As shown in Figure 7, all individual study results agreed with the overall results ($P < 0.001$): i-GSI was always the fastest, FSM-based GSI was the slowest, and APP-based GSI was slightly faster than PR-based GSI.

ST distribution of the same GSI among different individuals is studied. The results demonstrated that the GSI performance is individual dependent ($P < 0.01$), directly proportional to the operating ability of the subjects.

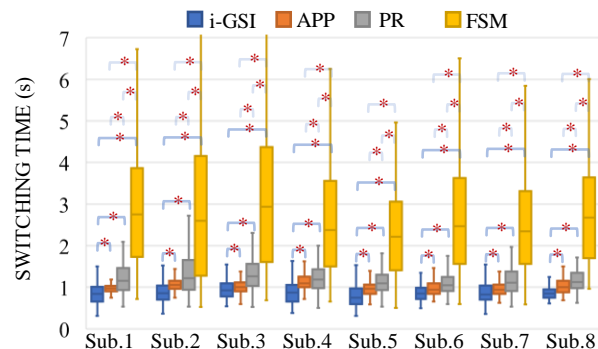

Fig. 7. Result in diff. Subjects. $P_* < 0.001$

*3) Grasp Types:* ST distribution of the same GSI among different grasp types was studied, and the results are shown in Figure 8. Only i-GSI had the same distribution ($P = 0.16$), so the ordering of grasp-type icons on the panel does not have a noticeable impact, and there is no need to optimize the sequence according to the frequency of use [32]. For APP-based GSI, *Tripod* and *Pinch* are distributed differently from the others ($P < 0.001$). In terms of ST, they were slower than the rest ($T_{Tripod}$, $_{Pinch} = 1.03$ s VS $T_{Others} = 0.97$ s). For PR-based GSI, only *Cylindrical* and *Spherical* (EMG pattern: *Wave In*, *Wave Out*) have the same distribution; all other grasp types are differently distributed ($P_{Tripod \text{ VS } Pinch} = 0.01$, $P_{Other\ pairs} < 0.01$). *Lateral* (EMG pattern: *double-tab*) is significantly slower than others. For FSM-based GSI, *Hook* is significantly different from others ($P_{Hook \text{ VS } Pinch}$, $_{Hook \text{ VS } Spherical}$, $_{Hook \text{ VS } Cylin.} < 0.01$, $P_{Hook \text{ VS } Tripod} = 0.018$). There is also a noticeable difference between *Lateral* and *Spherical* ($P_{Lateral \text{ VS } Spherical} = 0.048$). The difference was caused by the total NCCs necessary, especially the difference between *Hook* and other grasp types ($N_{Hook} = 795$, $N_{Lateral} = 864$, $N_{Cylindrical} = 893$, $N_{Pinch} = 914$, $N_{Tripod} = 908$, $N_{Spherical} = 974$). The total NCCs have a critical effect on ST.

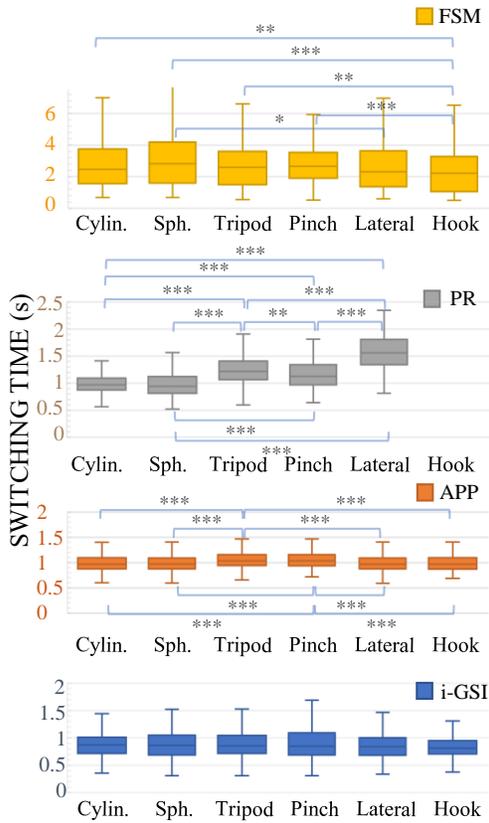

Fig. 8. Result in diff. grasp types. $0.01 < P_* < 0.05, 0.001 < P_{**} < 0.01, P_{***} < 0.001$

*4) Influence of NCCs on FSM-Based GSI:* Since ST of FSM-based GSI is sensitive to the total NCCs, a further study was conducted to analyze this sensitive relationship. As shown in Figure 9, as the NCCs increase, the median of ST grows proportionally ($y = 0.9036 x$), and the stability significantly worsens (the maximum values show a divergent trend). In other words, as the NCCs increase, ST increases proportionally, and controllability decreases.

All in all, through comprehensive comparison, i-GSI is the best method among the four. In the case such SSR of the four GSIs are similar, the PR-based GSI has the worst LE (81.5%) and a slow overall ST (1.14 s) which varies across different grasp types. The APP-based GSI has slightly better LE (90.0%) and a faster ST (1.00 s, when the screen is always-on), but it requires an additional mobile or tablet device and a healthy hand to assist. The FSM-based GSI has a poor LE (86.9%) and the slowest ST (2.53 s). The more NCCs required, the worse it performs. In contrast, i-GSI has the highest LE (93.5%) and the fastest ST (0.84 s). It has the same ST for all grasp types and can function well without the assistance of a healthy hand.

## V. DISCUSSIONS

Proficiency in the GSIs directly affects the experimental results. It is necessary to ensure that subjects conduct tryout-phase tests with proficiency. In the past, researchers judged whether subjects had mastered a GSI by subjective feeling, which is not rigorous. During the practice phase, subjects were frequently asked about mastery level, and most were found overoptimistic. Many believed they had mastered the GSI after memorizing it, but statistics show that the overall operation can still be immature even if the GSI has been well memorized. Using the BCG experience curve, the learning process of the experimental operations can be quantitatively reflected, and the learning period from novice to mastery can be estimated.

At present, affected by confounding factors such as load variation, forearm movement, and sweating, the effect of the traditional EMG-based GSIs perform far worse in practical applications than in ideal experimental conditions. The effect on patients is worse than on healthy subjects [14]. It is, therefore, necessary to test the applicability of i-GSI in real settings, especially on patients. A hypothesis was made that head motion is the dominant challenge in practical applications. Therefore, the subjects were instructed to change between two head poses (standby and tryout phases) during the experiment. Surprisingly, as a result, even if the head is in motion before viewing the object, i-GSI still achieves outstanding results. i-GSI was tested on a patient with congenital upper limb deficiency. Figure 10(a) shows that the patient demonstrated the same ST as five healthy subjects and faster ST than the other three. Figure 10(b) shows that the patient with 100% SSR achieved similar or better accuracy than all healthy subjects. So, i-GSI is also suitable for patients.

AR feedback is a powerful addition to i-GSI, effectively enhancing the SSR. The traditional implementation of FSM-based GSI (using LED feedback) and PR-based GSI (without feedback) cannot achieve such high SSRs as this paper. For FSM-based, there is insufficient feedback to effectively switch between six grasp-types if only LED are used, according to the subjects. For PR-based, as shown in Figure 11, the SSR without feedback (82.5%) is significantly lower than with feedback (99.5%). Another significant advantage of i-GSI is its error correction capability. In terms of ST, i-GSI only takes 1.06 s to correct one error, or 1.37 s for two errors, while PR-based GSI needs 1.75 s for a single error. For i-GSI, an error will be automatically corrected when the users are looking for the target grasp type. But for other GSIs, when the users find an error by visual feedback, they still need to perform additional actions (such as making a fist) to correct it.

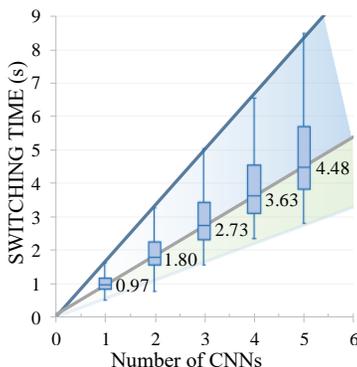

Fig. 9. Relation between ST and CCNs for FSM-based GSI

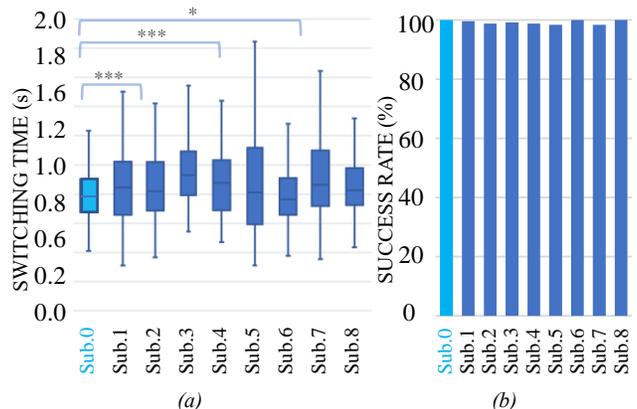

Fig. 10. Outcome of i-GSI on patient. Sub.0: patient, $0.01 < P_* < 0.05, P_{***}$

AR technology and eye-tracking technology can greatly expand the control bandwidth of HMI. Besides the grasp-type switching task described in this paper, artificial wrists, elbows, and other DOFs that require independent control can also be included in the panel as a "grasp-type icon." The combination of eye-tracking and AR can also improve feedback: AR displays additional information about the object which the eye movement indicates as a target of most interest to the user [28]. Furthermore, i-GSI runs as an independent module in HoloLens2 and can be designed as a Blackbox that may be easily integrated into the existing advanced prosthetic systems, which is the goal we wish to achieve next.

## VI. Conclusion

The prosthetic hand systems still lack an efficient and reliable interface for switching different grasp types. To address this issue, this paper proposed a new augmented reality and eye-tracking technology-based GSI: the i-GSI. Our experiments showed that, compared to three traditional GSIs (PR-based GSI, APP-based GSI, and FSM-based GSI), i-GSI demonstrated the utmost performance. In the future, we will integrate this method as an independent module into our existing prosthetic system and invite patients to carry out more verification tests in real-life settings.